\documentclass{article}

\PassOptionsToPackage{numbers, compress}{natbib}

\usepackage[preprint]{neurips_data_2024}

\usepackage[utf8]{inputenc} 
\usepackage[T1]{fontenc}    
\usepackage{hyperref}       
\usepackage{url}            
\usepackage{booktabs}       
\usepackage{amsfonts}       
\usepackage{nicefrac}       
\usepackage{microtype}      
\usepackage{xcolor}         

\newcommand{\remove}[1]{}

\usepackage{graphicx}
\usepackage{paralist}
\usepackage[normalem]{ulem}
\usepackage{amsmath}
\usepackage[ruled,vlined]{algorithm2e}

\usepackage{caption}
\usepackage{subcaption}

\title{CTBench: A Comprehensive Benchmark for Evaluating Language Model Capabilities in Clinical Trial Design}

\author{%
  Nafis Neehal\\
  \thanks{Github Link - \href{https://github.com/nafis-neehal/CTBench_LLM}{https://github.com/nafis-neehal/CTBench\_LLM}.}
  Department of Computer Science\\
  Rensselaer Polytechnic Institute\\
  Troy, NY 12180 \\
  \texttt{neehan@rpi.edu} \\
  \And
  Bowen Wang \\
  Center of Biotechnology and Interdisciplinary Studies  \\
  Rensselaer Polytechnic Institute \\
  Troy, NY 12180 \\
  \texttt{wangb19@rpi.edu} \\
  \And
  Shayom Debopadhaya \\
  Albany Medical College \\
  Albany, NY 12208\\
  \texttt{debopas@amc.edu} \\
  \And
  Soham Dan \\
  IBM Research \\
  1101 Kitchawan Rd, NY 10598 \\
  \texttt{soham.dan@ibm.com} \\
  \And
  Keerthiram Murugesan \\
  IBM Research \\
  1101 Kitchawan Rd, NY 10598 \\
  \texttt{keerthiram.murugesan@ibm.com} \\
  \And
  Vibha Anand \\
  Healthcare and Life Sciences, IBM Research \\
  314 Main Street, Cambridge, MA \\
  \texttt{anand@us.ibm.com} \\
  \And
  Kristin P. Bennett \\
  Department of Mathematical Sciences \\
  Rensselaer Polytechnic Institute \\
  Troy, NY 12180 \\
  \texttt{bennek@rpi.edu} \\
}

\begin{document}

\maketitle

\begin{abstract}

We introduce CTBench, a benchmark to assess language models (LMs) in aiding clinical study design. Given metadata specific to a study, CTBench examines how well AI models can determine the baseline features of the clinical trial (CT) which include demographic and relevant features collected at the start of the trial from all participants. The baseline features, typically presented in CT publications (often as Table 1), are crucial for characterizing study cohorts and validating results. Baseline features, including confounders and covariates,  are also required for accurate treatment effect estimation in studies involving observational data. CTBench consists of two datasets: "CT-Repo", containing baseline features from $1,690$ clinical trials sourced from \url{clinicaltrials.gov}, and "CT-Pub", a subset of $100$ trials with more comprehensive baseline features gathered from relevant publications. We develop two LM-based evaluation methods for evaluating the actual baseline feature lists against LM-generated responses. “ListMatch-LM” and “ListMatch-BERT”  use GPT-4o and BERT scores (at various thresholds), respectively, to perform the evaluation.  To establish baseline results,  we apply advanced prompt engineering techniques using LLaMa3-70B-Instruct and GPT-4o in zero-shot and three-shot learning settings to generate potential baseline features. We validate the performance of GPT-4o as an evaluator through human-in-the-loop evaluations on the CT-Pub dataset, where clinical experts confirm matches between actual and LM-generated features. Our results highlight a promising direction with significant potential for improvement, positioning CTBench as a useful tool for advancing research on AI in CT design and potentially enhancing the efficacy and robustness of CTs. 

\end{abstract}

\section{Introduction} 

Medical research can be broadly categorized into clinical trials (CTs) and observational studies, among other types. CTs aim to test one or more interventions for the improvement of health outcomes, where human subjects are recruited and assigned prospectively to the interventions or respective placebo controls. In contrast, observational studies are where the causal effects of health outcomes are observed by the investigators without controlling the independent variables. Randomized CT remains the “gold standard” in evaluating the safety and efficacy of the intervention. At the same time, observational studies allow for much less expensive and larger-scale investigations using existing or prospective data \cite{silverman2009randomized, faraoni2016randomized, thadhani2006formal}. In either case, it is crucial to ensure the balance between the study groups at the baseline, and that no systemic difference between study groups interferes with the causal relationship between the variables of interest and study outcomes \cite{roberts1999baseline}. Baseline characteristics, typically found in “Table 1” in CT publications, describe the demographic and relevant features collected at the beginning of the study for all participants between study groups. Depending on the study outcomes, the baseline characteristics may include sociodemographics, anthropometrics, confounding medical conditions, etc. For observational studies, the baseline features can help design the study by matching the cohort by the confounders and covariates. The showcase of baseline characteristics shows the reader how representative the study population is and how applicable the results would be. It validates the study design, increases the statistical efficiency, and helps the investigators draw logical conclusions \cite{holmberg2022adjustment, festic2016improve, zhang2017comparing}. 

Currently, general guidelines and considerations for the selection of baseline features exist \cite{prsinfoClinicaltrials}. However, most of the relevant features are study-specific and require the investigators’ judgment. This may lead to an overlook of relevant confounders or covariates. Alternatively, for observational studies in particular, the improper selection of confounders/covariates from baseline features may lead to over-adjustment bias \cite{van2024overadjustment}. In addition, the reporting of baseline feature variables is not standardized and consistent across studies even for similar interventions or health outcomes. To tackle this issue in clinical research, we introduce CTBench, a benchmark to assess the role of language models (LMs) in aiding clinical study design. CTBench requires these models to predict the baseline characteristic variables of a clinical study based on the CT metadata. This study is the first to use LMs to solve the challenging task of designing the baseline features for both CTs and observational studies. 

To achieve this, we create the benchmark from the centralized CT repository along with human annotation. We create two expansive datasets: \begin{inparaenum}[1)]
    \item “CT-Pub” which includes the metadata and baseline features from 1,690 CTs collected from the \url{clinicaltrials.gov} API, and,
    \item “CT-Repo” which contains a subset of 100 trials where the baseline features are retrieved from the related clinical publications via human curation. 
\end{inparaenum}  

The main contributions of this work include: \begin{inparaenum}[1)]
    \item we propose a benchmark (CTBench) to use LMs to develop AI support tools for CT, assist researchers in selecting baseline features and design more efficient and robust clinical studies;   
    \item we create two CT metadata datasets with associated baseline features derived from a definitive repository and published papers; 
    \item we develop two automated evaluation methods for comparing predicted and actual trial baseline features,   ``ListMatch-LLM" and ``ListMatch-BERT", and validate them with “human-in-the-loop” evaluations; and 
    \item we demonstrate CTBench by using robust prompt engineering techniques on several LLMs to generate the baseline feature variables and evaluate their performance results. 
    \item Our data, code, and demo examples are available at \href{https://github.com/nafis-neehal/CTBench_LLM}{https://github.com/nafis-neehal/CTBench\_LLM}.
\end{inparaenum}  

\section{Related Work}%

Recent applications of LLMs show that they can serve as powerful tools alongside human evaluators \cite{hutson2024ai, ghim2023transforming}. They have been efficiently deployed for extracting clinical information with models such as the CT-BERT and MT-clinical BERT \cite{liu2021clinical, mulyar2021mt}. CliniDigest showed a similar value, reducing 10,000-word CT descriptions into 200-word summaries using GPT 3.5 \cite{white2023clinidigest}. LLMs have been shown to have further uses in comparing similarity among trials to improve result comparison and aid in the precision design of subsequent studies \cite{wang2022trial2vec}. Advances in prompting have additionally increased the use cases, both in specific medical specialties and generalized contexts \cite{wang2023autotrial, lee2024seetrials, singhal2023large}

Research exists on using LLMs to aid in creating eligibility criteria for CTs \cite{yuan2019criteria2query, jin2023matching, datta2024autocriteria, hamer2023improving}. Critical2Query was validated on 10 CTs of different medical contexts to produce inclusion and exclusion criteria for the resolution of previous conditions, disease severity, and disease duration \cite{yuan2019criteria2query}. TrialGPT proposed an LLM that could potentially reduce 42.6\% of the screen time needed to match CTs by domain experts without compromising in near-expert level grouping \cite{jin2023matching}. AutoCriteria similarly shows promising extraction of eligibility criteria through a set of 180 manually annotated trials \cite{datta2024autocriteria}. 

However, automation of proposing baseline features of CTs is lacking. Since baseline features of CTs have become significantly more complex from 2011-2022 \cite{markey2024clinical}, better approaches for suggesting a generalizable and standardized set of cohort demographics and features are needed. Adequately training and validating LLMs for these clinical tasks requires relevant and feature-rich datasets. Several works have leveraged the \url{clinicaltrials.gov} database that has information for over 300,000 research studies conducted in more than 200 countries \cite{liu2021clinical, wang2022trial2vec, wang2023autotrial, yuan2019criteria2query, datta2024autocriteria}. However, the prioritization of creating CT eligibility tools has left patient descriptor data relatively understudied.

CTBench addresses gaps between study criteria and features that are reported in databases such as \url{clinicaltrials.gov} in comparison to what appears in the final publication. For example, where age, sex, race, ethnicity, region of enrollment, and hemoglobin A1C may be reported on databases \cite{NCT03987919Portal}, investigators ensured that additional baseline characteristics of fasting serum glucose, duration of diabetes, BMI, weight, waist circumference, estimated GFR, albumin-to-creatinine ratio, medication use, and cardiovascular parameters were included in the final report \cite{frias2021tirzepatide}. As only 4 baseline features are consistently reported by greater than 10\% of studies on these well-used databases, the development of publicly available and accurate baseline feature databases is necessary \cite{cahan2017second}. Current datasets that attempt to address this are limited by low CT cohort size or have sufficient patient data but are sourced from general clinical notes in place of CTs \cite{koopman2016test, roberts2021overview}. Other projects do create datasets from high-quality, manually annotated CTs, but do not provide public access \cite{datta2024autocriteria}. Here, our constructed datasets are relevant to baseline demographics (CT-Repo, CT-Pub), with human annotation to include all the features of a reported clinical study (CT-Pub), and larger than previously available CT data sets with a complete set of patient demographic data \cite{koopman2016test, roberts2021overview}.

\section{Methodology}
\label{sec:methodology}
\subsection{Data Construction}
\begin{table}[t]
\centering
\caption{A sample example from CTBench with CT metadata and corresponding baseline features.}
\label{tab:trial-metadata}
\resizebox{\textwidth}{!}{%
\begin{tabular}{@{}lll@{}}
\toprule
\textbf{Field} & \textbf{Data}                                                                                                                                          &  \\ \midrule
\textbf{Trial ID}      & NCT00000620                                                                                                                                   &  \\ \\
\textbf{Trial Title}   & Action to Control Cardiovascular Risk in Diabetes (ACCORD)                                                                                    &  \\ \\
\textbf{Brief Summary} &
  \begin{tabular}[c]{@{}l@{}}The purpose of this study is to prevent major cardiovascular events (heart attack, stroke, or cardiovascular death) in adults \\ with type 2 diabetes mellitus using intensive glycemic control, intensive blood pressure control, and multiple lipid management.\end{tabular} &
   \\ \\
\textbf{Eligibility Criteria} &
  \begin{tabular}[c]{@{}l@{}} \textit{Inclusion Criteria:}\\ * Diagnosed with type 2 diabetes mellitus, as determined by the new American Diabetes Association guidelines, \\ which include a fasting plasma glucose level greater than 126 mg/dl (7.0 mmol/l), or a 2-hour postload value in the \\ oral glucose tolerance test of greater than 200 mg/dl, with confirmation by a retest\\ ....\\ \textit{Exclusion Criteria:}\\ ...\end{tabular} &
   \\ \\
\textbf{Conditions}    & Atherosclerosis, Cardiovascular Diseases, Hypercholesterolemia, ... &  \\ \\
\textbf{Primary Outcomes} &
  \begin{tabular}[c]{@{}l@{}}First Occurrence of a Major Cardiovascular Event (MCE), ... \end{tabular} &
   \\ \\
\textbf{Interventions} & Anti-hyperglycemic Agents, Anti-hypertensive Agents, ...      &  \\ \\ 
\textbf{\begin{tabular}[c]{@{}l@{}}Baseline Features \end{tabular}} &
  \begin{tabular}[c]{@{}l@{}}Age, Gender, Ethnicity (NIH/OMB), Race, Region of Enrollment, Previous cardiovascular disease (CVD) event, \\ Glycated hemoglobin, Blood pressure, Cholesterol, Triglycerides, Diabetes duration\end{tabular} &
   \\ \bottomrule
\end{tabular}%
}
\end{table}

We collect CT data from \url{clinicaltrials. gov} using their publicly available API. Our selection criteria include studies that are: \begin{inparaenum}[1)] \item interventional trials, \item completed with results reported, \item related to one of five common chronic diseases: hypertension, chronic kidney disease, obesity, cancer, diabetes, and \item reported at least six baseline features. \end{inparaenum} The requirement for a minimum of six baseline features ensures the inclusion of studies with more comprehensive data beyond commonly reported features such as age group, race/ethnicity, and sex. This criterion is implemented to ensure the robustness of our dataset, as some features from the publication about CT may not be reported on the \url{clinicaltrials. gov}.

For each CT, we collect several types of information (see Table \ref{tab:trial-metadata}). We initially started with 1798 studies returned from the API query. After thorough pre-processing steps, including removing duplicate trials and trials with missing values, we are left with 1693 CTs for our final study.

From our 1693 CTs, we construct two datasets: "CT-Repo" and "CT-Pub" summarized in Table \ref{tab:data-stat} The CT-Repo dataset consists of 1690 trials, with the remaining three trials used as example trials for three-shot learning in LMs. We randomly pick 100 CTs from the CT-Repo dataset to build the CT-Pub dataset.  For each trial in CT-Pub, human annotators manually collect the list of baseline features reported in the publications associated with the CT and ensure that: \begin{inparaenum}[1)] \item each CT has at least one relevant publication reporting the trial results, \item the publication contains a table where the baseline features featured for the trial are fully reported, and \item the publication is evidenced to be connected to the trial by mentioning the trial ID in the publication and/or in the publisher's website. \end{inparaenum} 

\begin{table}[!ht]
\centering
\caption{Dataset description for CTBench.}
\label{tab:data-stat}
\resizebox{\textwidth}{!}{%
\begin{tabular}{@{}lcccccc@{}}
\toprule
 &
  \begin{tabular}[c]{@{}c@{}}Total\\ n\end{tabular} &
  \begin{tabular}[c]{@{}c@{}}Cancer \\ n (\%)\end{tabular} &
  \begin{tabular}[c]{@{}c@{}}Chronic Kidney Disease \\ n (\%)\end{tabular} &
  \begin{tabular}[c]{@{}c@{}}Diabetes \\ n (\%)\end{tabular} &
  \begin{tabular}[c]{@{}c@{}}Hypertension \\ n (\%)\end{tabular} &
  \begin{tabular}[c]{@{}c@{}}Obsesity \\ n (\%)\end{tabular} \\ \midrule
CT-Repo &
  1690 &
  484 (28.64\%) &
  169 ( 10.00\%) &
  479 (28.34\%) &
  266 (15.74\%) &
  292 (17.27\%) \\
CT-Pub &
  100 &
  16 (16.00\%) &
  18 (18.00\%) &
  34 (34.00\%) &
  14 (14.00\%) &
  18 (18.00\%) \\ \bottomrule
\end{tabular}%
}
\end{table}

\textbf{Challenges:}
\label{sec:data_challenges}
The data extracted from \url{clinicaltrials.gov} include title, summary, conditions, eligibility criteria, interventions, primary outcomes, and baseline features in free-text format (Table \ref{tab:trial-metadata}). The trial titles and brief summaries provide an overview of the study in plain language, often without consistent terminology. 
Conditions refer to health issues/diseases being studied written in free text, which can lead to inconsistencies in interpretation due to polysemy (multiple meanings) and synonymy (different terms for the same concept). Eligibility criteria, encompassing both inclusion and exclusion criteria, are detailed as paragraphs, bulleted lists, or enumeration lists, without adherence to common standards or controlled vocabularies. Interventions describe the treatments or procedures being tested, in unstructured text. Primary outcomes and baseline features outline the main objectives and initial data points of the study, respectively, and are similarly unstructured, lacking standardization in terms of medical dictionaries or ontologies. This variability and lack of standardized language across all these fields pose significant challenges for both data extraction and results analysis.

\subsection{Generation Task}
\begin{figure}[t]
    \centering
    \includegraphics[width=\linewidth]{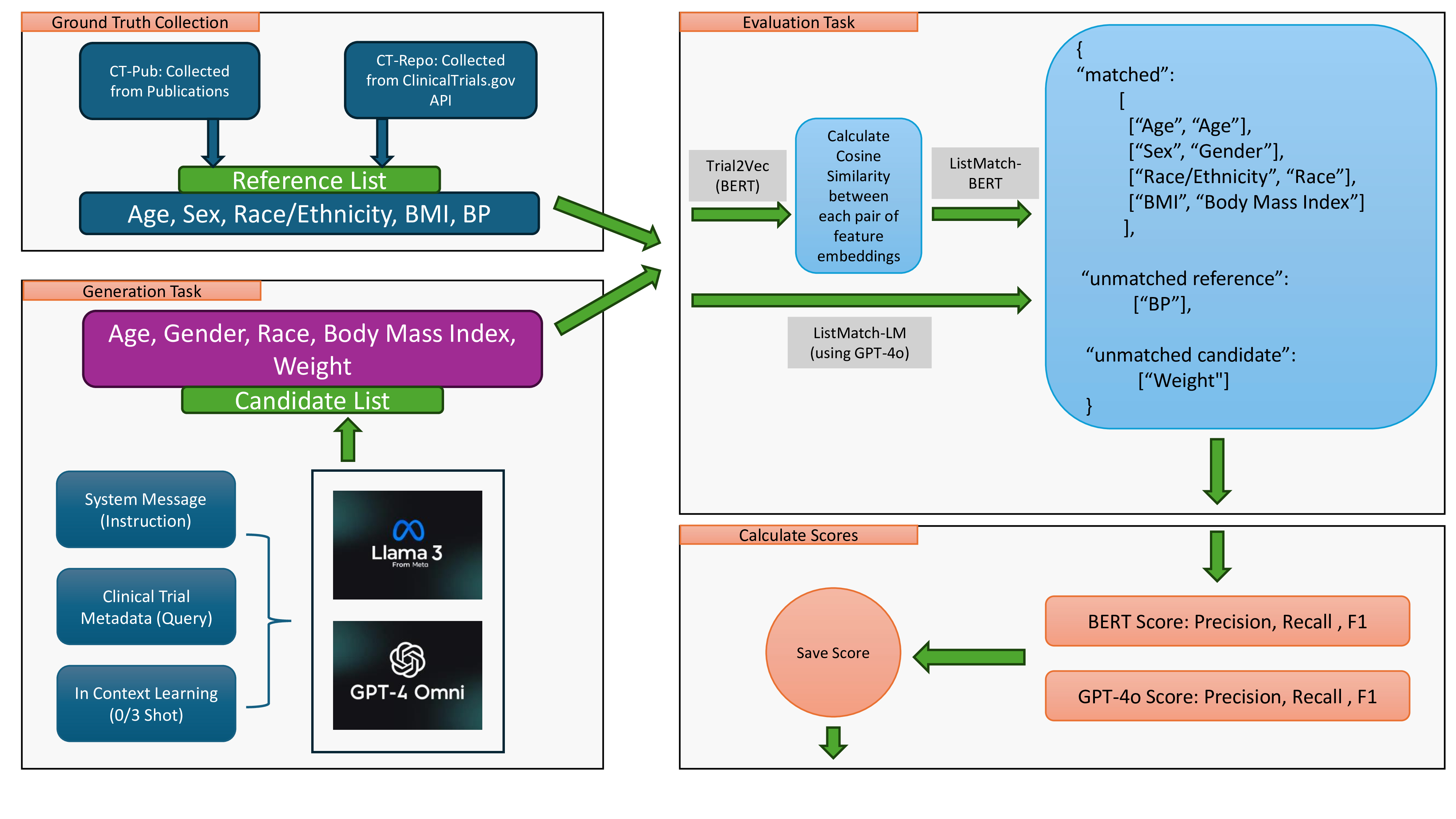}
    \caption{Workflow of CTBench.}
    \label{fig:gen_eval_process}
\end{figure}

The CTBench task is to predict the baseline features of a study given the metadata.  We demonstrate our benchmarking process and evaluate performance results on two state-of-the-art LMs, 
open-source LLaMa3-70B-Instruct \cite{llama3modelcard} and commercial GPT-4o \cite{openaiGPT4}. For GPT-4o, we used the API provided by OpenAI \cite{openaiAPI}. For LLaMa-3-70B-Instruct, we used APIs from GROQ \cite{groqGroqBuilds} and HuggingFace's serverless inference service \cite{huggingfaceInferencePROs}.  We investigate two in-context learning settings for feature generation: zero-shot and three-shot \cite{dong2022survey}. Each query has the system message and the user query (Figure \ref{fig:gen_eval_process}). For the zero-shot setting, we provide CT metadata (excluding the baseline features) as input context to these models (Figure \ref{fig:gen-eval-prompt}), and query the models to generate a list of probable baseline features relevant to the clinical trial. In the three-shot setting (see Appendix C for full prompt template), we extend the zero-shot system prompt by appending trial metadata and corresponding answers (i.e., list of baseline features) for three example trials. All our generation prompts are in Appendix C. For CT-Repo, the generation task involves predicting the list of baseline features reported in the \url{clinicaltrials.gov} portal using the CT metadata presented in Table \ref{tab:trial-metadata}. For the CT-Pub dataset, the generation task is to predict the baseline features collected from the publications relevant to each trial. 

\subsection{Evaluation Task}
\label{subsec:eval_task}
 The evaluation task compares the "candidate features" suggested by each LLM with the "reference baseline features" from the CT publications for CT-Pub or \url{clinicaltrials.gov} API for CT-Repo. 
The objective is to evaluate each pair of features, one from the reference list and one from the candidate list, to determine if they are contextually and semantically similar, i.e., if they match. We remove noisy keywords from the feature lists (e.g., "Customized," "Continuous") during pre-processing. 
After identifying all matched pairs, the final results are categorized into three lists: matched pairs, unmatched reference features, and unmatched candidate features. 
We employ two approaches for identifying matched pairs: ``ListMatch-BERT" and ``ListMatch-LM." 
For the evaluation task, we use Trial2Vec and GPT-4o for ListMatch-BERT and ListMatch-LM, respectively. The Trial2Vec implementation requires local installation and a GPU for inference, as it is not readily available through HuggingFace or other inference service providers. We utilized NVIDIA Ampere A100 and NVIDIA T4 GPUs via Google Colab for our work. For GPT-4o as an evaluator, we again used the OpenAI APIs available through their public site.  All hyperparameters related to our generation and evaluation tasks are presented in Appendix B. We use a fixed seed and a temperature value of 0.0 across all experiments to ensure the outputs are deterministic and reproducible \cite{openaiReproduce}. 

\begin{figure}[t]
    \centering
    \includegraphics[width=\linewidth]{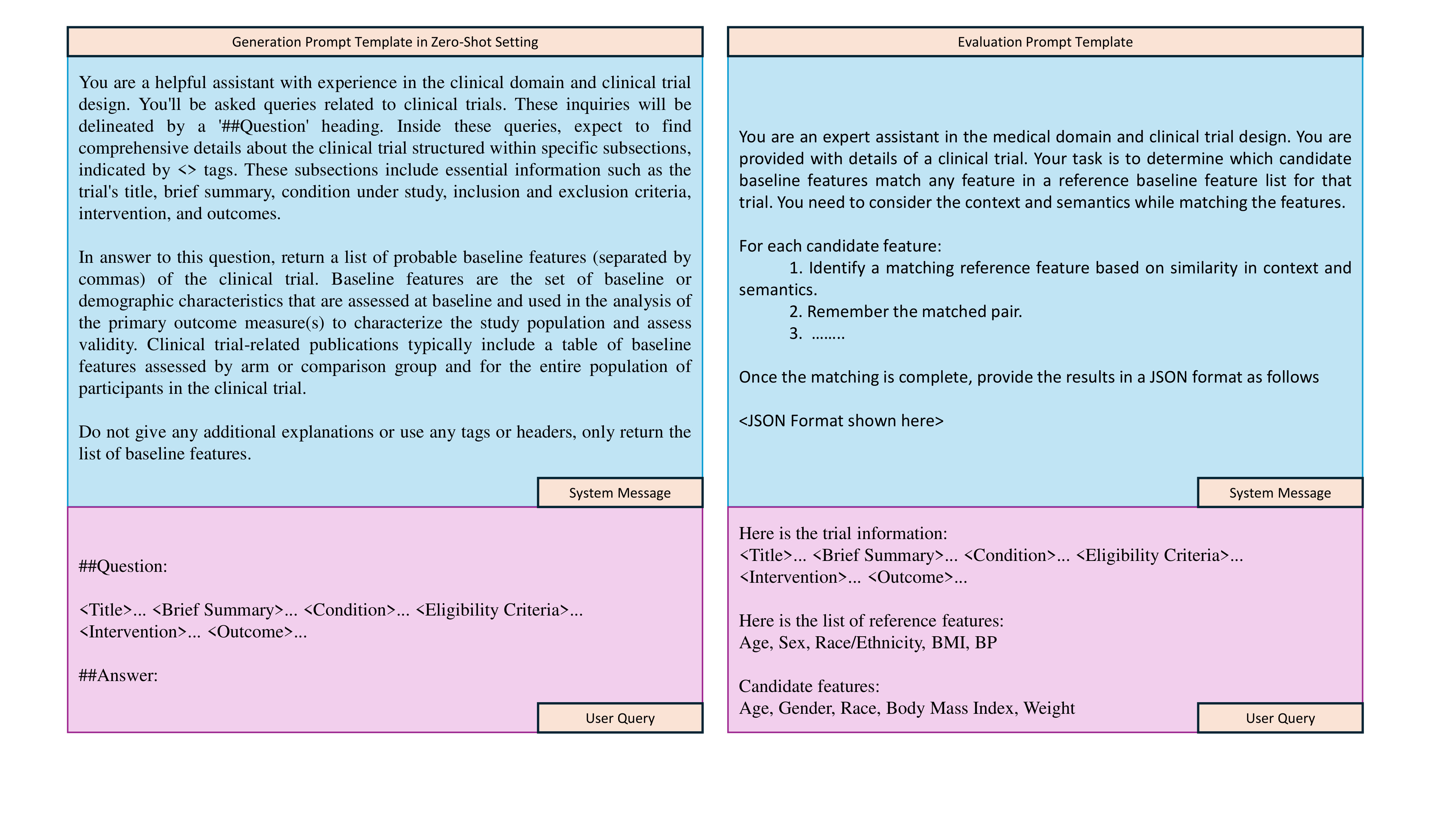}
    \caption{Prompt template for generation (in zero-shot setting) and evaluation}
    \label{fig:gen-eval-prompt}
\end{figure}

\textbf{ListMatch-BERT:} Here we consider a variation of the BERTScore \cite{zhang2019bertscore}. We utilize Trial2Vec architecture proposed for CTs, built on top of TrialBERT \cite{wang2022trial2vec} (MIT license) to generate embeddings for each feature and then calculate a cosine similarity matrix for each set of pairs. We explore different matching threshold values $T_h \in \{0.6, 0.7, 0.8, 0.9\}$, and recommend using the value of 0.7 (see Appendix D for detailed comparison and reasoning). Matches are considered starting from the pair with the highest cosine similarity above $T_h$, and these pairs are added to the matched list, and removed from their respective lists and the similarity matrix. Matching continues until: \begin{inparaenum}[1)] \item no more matches are found with similarity greater than $T_h$, or \item no more features remain to match in either the reference or candidate list. \end{inparaenum} A detailed description of the ListMatch-BERT process is provided in Appendix A. 

We report mean Precision, mean Recall, and mean F1 scores across all studies for each dataset.  Once the lists of matched pairs, unmatched references, and unmatched candidates are established, and given: \(\text{TP}\) (True Positives): \(n_{matched\_pairs}\), \(\text{FP}\) (False Positives): \(n_{remaining\_candidate\_features}\), \(\text{FN}\) (False Negatives): \(n_{remaining\_reference\_features}\), we calculate precision and recall: 

\begin{equation}
\text{Precision} = \frac{\text{TP}}{\text{TP} + \text{FP}} = \frac{n_{matched\_pairs}}{n_{matched\_pairs} + n_{remaining\_candidate\_features}}
\end{equation}

\begin{equation}
\text{Recall} = \frac{\text{TP}}{\text{TP} + \text{FN}} = \frac{n_{matched\_pairs}}{n_{matched\_pairs} + n_{remaining\_reference\_features}}
\end{equation}


\textbf{ListMatch-LM:} Here GPT-4o is prompted to identify matched pairs and the remaining unmatched sets (see Figures \ref{fig:gen_eval_process} and \ref{fig:gen-eval-prompt}). For each study, GPT-4o receives the reference features and candidate features as input. Trial metadata (excluding the actual baseline features) is provided as context. GPT-4o is tasked with identifying matched pairs and generating unmatched lists, which are returned as a JSON object. Mirroring the procedure used in ListMatch-BERT, the model is instructed to remove matched pairs from further consideration immediately upon identification, ensuring that no reference feature is matched to multiple candidate features, and vice versa. Once the matches are generated and the unmatched items are identified, we calculate precision, recall, and F1 scores similarly as described above and report their means over all the studies. Appendix C provides the full evaluation prompt.

\textbf{Human Evaluation:} To evaluate the accuracy of GPT-4o as an evaluator, we employ clinical domain experts to serve as human annotators. Their task is to identify matched pairs for each of the 100 CT studies in the CT-Pub dataset. To streamline the evaluation, we focus exclusively on the candidate responses generated by GPT-4o in the three-shot setting. The annotators receive the same information provided to GPT-4o during its evaluation and are instructed to match features using the same criteria. We developed a web tool to collect and store the responses from all annotators for each of the 100 studies in a database.  We also solicit evaluations from human annotators regarding the remaining unmatched candidate features that may merit further examination. Our findings indicate a high level of agreement between the human annotator and GPT-4 Omni's evaluations, underscoring the reliability of GPT-4o in capturing nuanced similarities between features. Detailed results of these experiments are provided in Appendix D.  

\section{Results and Discussion}

\begin{figure}[t]
    \centering
    \includegraphics[width=\linewidth]{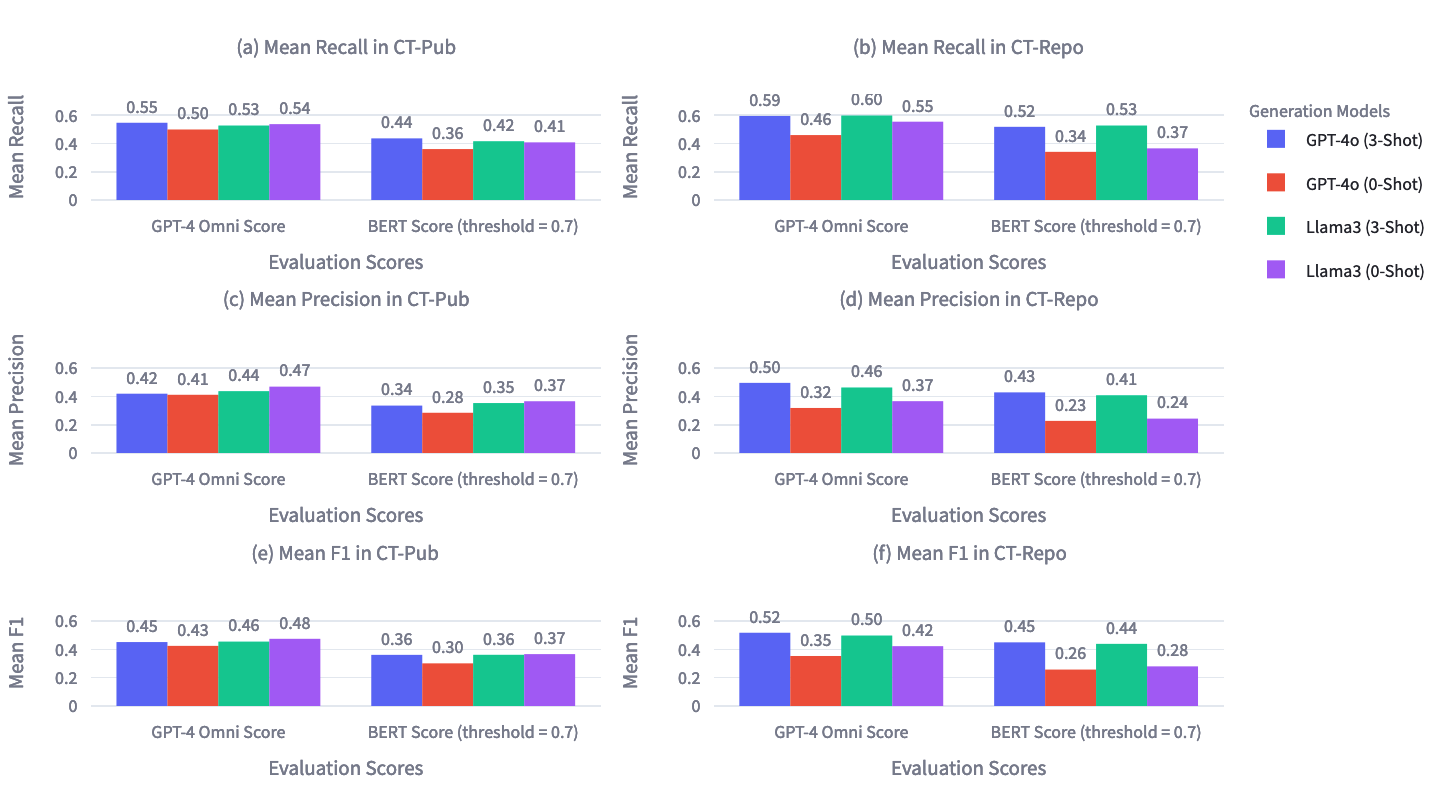}
    \caption{Performance Comparison for CT-Pub and CT-Repo datasets}
    \label{fig:combined_plots}
\end{figure}

In CTBench, precision measures the proportion of predicted baseline features that are accurate, while recall measures the proportion of actual baseline features that the model successfully identifies. We find recall to be of more interest as it ensures comprehensive identification of all relevant baseline features, which is crucial for accurately characterizing study cohorts and maintaining the validity and robustness of clinical trial results. High recall minimizes the risk of missing critical features that could undermine the study's conclusions. Figure \ref{fig:combined_plots} shows the performance comparison of GPT-4o and LLaMa3 for CT-Pub and CT-Repo datasets. We find that GPT-4o (3-Shot) leads in recall in the CT-Pub dataset, while LLaMa3 (0-Shot) excels in the CT-Pub dataset for precision and F1 scores. In the CT-Repo dataset, GPT-4o (3-shot) outperforms LLaMa3 across all ICL settings and metrics. 

\subsection{Performance Analysis in Generation Tasks}

\subsubsection{Analysis on CT-Pub Dataset} 

\textbf{Observation about Metric Values and Model Performance:} The values of recall, precision, and F1 scores are not particularly high, indicating a moderate performance of LLaMa3 and GPT-4o in predicting baseline features. This suggests there is room for improvement in the models' ability to generate accurate and comprehensive baseline features.

\textbf{Comparison of Precision, Recall, and F1 Scores Across Models:} The models exhibit varied strengths across different metrics. LLaMa3 (0-Shot) demonstrates the highest precision and F1 score, with an F1 score of 0.48, indicating its strong capability to accurately identify relevant baseline features without requiring prior examples. GPT-4o (3-Shot) leads in the recall, highlighting its superior ability to retrieve a comprehensive list of relevant baseline features when examples are provided. This suggests that GPT-4o benefits significantly from example-based learning, whereas LLaMa3 performs robustly even in a zero-shot setting, making it a versatile choice for scenarios with limited training data.

\textbf{ICL Setting Analysis:}
\begin{itemize}
    \item \textbf{Zero-shot vs. Three-shot:} In the CT-Pub dataset, LLaMa3 performs better in the zero-shot setting, particularly in precision and F1 score. GPT-4o, however, benefits more from the examples, performing better in the three-shot setting in the recall. 
    \item \textbf{Model Benefit from Examples:} GPT-4o shows a significant improvement in recall when examples are provided (3-shot), whereas LLaMa3 shows a higher overall performance in the zero-shot setting.
\end{itemize}

\subsubsection{Analysis on CT-Repo Dataset:}
\textbf{Observation about Metric Values and Model Performance:}
Similar to the CT-Pub dataset, the values are not exceptionally high, reflecting moderate performance in predicting baseline features. This emphasizes the need for enhanced models to improve prediction accuracy and comprehensiveness.

\textbf{Comparison of Precision, Recall, and F1 Scores Across Models:}
The CT-Repo dataset reveals that GPT-4o (3-Shot) outperforms LLaMa3 in precision and F1 score, achieving a notable F1 score of 0.52, while providing comparable performance in recall. This highlights GPT-4o's robustness and effectiveness when prior examples are available, making it highly suitable for matching or adjusting treatment and control subjects in clinical trials and observational studies. LLaMa3 (3-Shot) also demonstrates strong performance, particularly in the recall, indicating its capability to retrieve a broad range of relevant features when examples are provided. The overall moderate performance of both models reflects the complexity and challenging nature of accurately predicting baseline features from clinical trial metadata. 

\textbf{ICL Setting Analysis:}
\begin{itemize}
    \item \textbf{Zero-shot vs. Three-shot:} In the CT-Repo dataset, both models perform better in the three-shot setting. GPT-4o significantly benefits from examples, especially in precision and recall.
    \item \textbf{Model Benefit from Examples:} GPT-4o shows substantial improvement with examples (3-shot), indicating its dependency on context for better performance. LLaMa3 also shows improved performance with examples but retains good performance in the zero-shot setting. Since the ground-truth baseline features for CT-Repo were collected from the \url{clinicaltrials.gov} API, there are specific nuances, such as reporting 'Region of Enrollment' as a baseline feature, which is not typically seen in CT-Pub publications. We believe this context explains why both GPT-4o and LLaMa3 benefit from example-based learning in this scenario.
\end{itemize}

\subsubsection{Why is GPT-4o under-performing significantly and consistently in zero-shot setting?} GPT-4o (zero-shot) underperforms across all cases and scores in both datasets due to the lack of contextual learning from prior examples, which is crucial for accurately interpreting and predicting complex, domain-specific clinical trial features. This setting relies solely on pre-trained knowledge, which is insufficient for the nuanced and detailed task of baseline feature prediction in clinical trials.

\subsection{Performance on Evaluation Tasks}
\textbf{GPT-4 Omni Scores:}
GPT-4 evaluation scores generally surpass BERT scores at a 0.7 threshold due to GPT-4o's broader understanding and contextual evaluation, which captures more nuanced similarities between reference and candidate baseline features. This results in a more generous and context-aware assessment compared to the stricter, more literal BERT scoring.

\textbf{BERT Scores (threshold = 0.7):}
After examining several thresholds, we recommend 0.7 to be used as the threshold value for producing BERT scores using ListMatch-BERT. The 0.7 threshold for BERT scores signifies a balance between generous and strict evaluation criteria, requiring high similarity for matches to be considered valid. This, however, reduces precision and recall by demanding closer alignment between generated and actual features compared to lower threshold values. Lowering the threshold would allow for more matches but could increase false positives and false negatives, affecting the precision and recall negatively. We present a thorough evaluation of BERT scores at different threshold values in Appendix D.

Comparing both metrics, we believe that GPT-4 Omni scores suggest a comprehensive and context-sensitive evaluation, crucial for accurately assessing the quality of LM-generated baseline features in clinical trial design.

\section{Limitations}
\label{sec:limitations}
\textbf{CT Data Expansion:} Our results, derived from CT data, demonstrate the potential of LLMs to significantly aid in the design and implementation of clinical studies. But the CTBench consists of only RCTs for 5 chronic diseases gathered from \url{clinicaltrials.gov} with only a subset annotated with additional "gold-standard" from CT-related papers.  Using our tools and framework, CTBench could be expanded with other CT repositories, more published CT results, and more diseases.  Future work should also explicitly incorporate and evaluate observational studies.  

\textbf{Evaluation Methods:} We have presented two LLM-based matching methods and associated evaluation metrics, but how to best evaluate predicted descriptors is an interesting research question in itself. Currently, each reference or candidate item is permitted to be matched only once to provide a standardized fair evaluation across models.   But other strategies allowing multiple matches are possible.  We hope that the human-in-the-loop evaluation tools provided to compare the LM and human evaluations assist in the further evolution of effective evaluation strategies.  

\textbf{Additional Methods for Generation:} Our baseline CTBench study focuses on benchmarking the two state-of-the-art LLaMa3-70B-Instruct and GPT-4o models only with zero-shot and three-shot prompts due to resource constraints.  By contrasting an open-source model (LLaMa3-70B-Instruct) with a closed-source model (GPT-4o), we aim to provide a preliminary evaluation of current leading technologies.  In our experiments, both for the text generation and evaluation API calls, we have maintained a consistent approach by using a fixed seed and a temperature value set to 0.0. This methodological choice is based on OpenAI's documentation \cite{openaiReproduce}, which claims that a fixed seed and a temperature parameter of 0.0 are likely to produce reproducible and deterministic results. But many other possibilities exist.  Running each API call multiple times with the same question and considering aggregated answers could improve results. We hope that CT-bench will spur new prompt and model research to expand the scope and depth of AI methods for CT design support.

\textbf{Impact of Societal Bias:} Societal biases present in language models (LMs) can potentially be transferred to clinical trials through the models' baseline feature predictions. This bias could skew the characterization of study cohorts, leading to biased clinical results and affecting the generalizability and applicability of the findings. Such biases in baseline features can undermine the validity of clinical trials, resulting in health outcomes that do not accurately reflect the broader population.

\section{Conclusion}
CTBench serves as a pioneering benchmark for evaluating LLMs in predicting baseline features from CT metadata - a critical component in CT design. By leveraging datasets from \url{clinicaltrials.gov} and curated from trial publications, and utilizing advanced evaluation methods such as ListMatch-LM and ListMatch-BERT, CTBench provides a robust framework for assessing AI-generated baseline features. Our results establish a promising baseline, validated through expert human evaluations, and underscore CTBench's potential to significantly enhance the efficacy and robustness of clinical trials through advanced AI research.

\newpage
\ack 

This work was supported by IBM Research and the Rensselaer Institute for Data Exploration and Applications. 

\medskip
\bibliography{references}



\section*{Checklist}


\begin{enumerate}

\item For all authors...
\begin{enumerate}
  \item Do the main claims made in the abstract and introduction accurately reflect the paper's contributions and scope?
    \answerYes{see section \ref{sec:methodology} - \ref{sec:limitations}}
  \item Did you describe the limitations of your work?
    \answerYes{see section \ref{sec:limitations}}
  \item Did you discuss any potential negative societal impacts of your work?
    \answerYes{see section \ref{sec:limitations}}
  \item Have you read the ethics review guidelines and ensured that your paper conforms to them?
    \answerYes{}
\end{enumerate}

\item If you are including theoretical results...
\begin{enumerate}
  \item Did you state the full set of assumptions of all theoretical results?
    \answerNA{}
	\item Did you include complete proofs of all theoretical results?
    \answerNA{}
\end{enumerate}

\item If you ran experiments (e.g. for benchmarks)...
\begin{enumerate}
  \item Did you include the code, data, and instructions needed to reproduce the main experimental results (either in the supplemental material or as a URL)?
    \answerYes{}
  \item Did you specify all the training details (e.g., data splits, hyperparameters, how they were chosen)?
    \answerYes{see section \ref{sec:methodology} and Appendix}
	\item Did you report error bars (e.g., with respect to the random seed after running experiments multiple times)?
    \answerNA{see section \ref{sec:limitations}}
	\item Did you include the total amount of compute and the type of resources used (e.g., type of GPUs, internal cluster, or cloud provider)?
    \answerYes{see Appendix}
\end{enumerate}

\item If you are using existing assets (e.g., code, data, models) or curating/releasing new assets...
\begin{enumerate}
  \item If your work uses existing assets, did you cite the creators?
    \answerYes{see section \ref{subsec:eval_task}}
  \item Did you mention the license of the assets?
    \answerYes{see \ref{subsec:eval_task}}
  \item Did you include any new assets either in the supplemental material or as a URL?
    \answerYes{see Github link + Appendix}
  \item Did you discuss whether and how consent was obtained from people whose data you're using/curating?
    \answerNA{}
  \item Did you discuss whether the data you are using/curating contains personally identifiable information or offensive content?
    \answerNA{}
\end{enumerate}

\item If you used crowdsourcing or conducted research with human subjects...
\begin{enumerate}
  \item Did you include the full text of instructions given to participants and screenshots, if applicable?
    \answerNA{}
  \item Did you describe any potential participant risks, with links to Institutional Review Board (IRB) approvals, if applicable?
    \answerNA{}
  \item Did you include the estimated hourly wage paid to participants and the total amount spent on participant compensation?
    \answerNA{}
\end{enumerate}

\end{enumerate}


\newpage
\appendix

\begin{center}
   {\huge{Appendix}} 
\end{center}

\section{ListMatch-BERT}
\label{app:listmatch-bert}

In this study, we examine a modified version of BERTScore. We leverage the Trial2Vec architecture, which is an extension of TrialBERT. This architecture generates embeddings for each feature, after which we compute a cosine similarity matrix for each pair of sets. We experiment with various matching threshold values $T_h \in \{0.6, 0.7, 0.8, 0.9\}$, ultimately recommending a threshold of 0.7 (refer to Appendix D for a comprehensive comparison and justification). Matching begins with the pair exhibiting the highest cosine similarity above $T_h$. These pairs are subsequently added to the matched list and removed from their respective lists and the similarity matrix. This process repeats until: \begin{inparaenum}[1)] \item no further matches surpass the similarity threshold $T_h$, or \item no remaining features exist to match in either the reference or candidate list. \end{inparaenum}
Here is the a detailed version of the ListMatch-BERT algorithm (Note: this Preprint Version has formatting and syntactical changes for ArXiv submission) -

\begin{algorithm}[H]
\caption{ListMatch-BERT}
\label{alg:get_match_json}
\SetKwInOut{Input}{Input}
\SetKwInOut{Output}{Output}

\Input{Similarity matrix $\mathbf{S}$, reference features $\mathbf{R}$, candidate features $\mathbf{C}$, threshold $T_h$}
\Output{Matched features, remaining reference features, remaining candidate features}

$\mathbf{M} \leftarrow \emptyset$\;
$\mathbf{R}_{\text{remaining}} \leftarrow \mathbf{R}$\;
$\mathbf{C}_{\text{remaining}} \leftarrow \mathbf{C}$\;
Iteration $\leftarrow 1$\;

\While{$\mathbf{S}$ has rows and columns}{
    Find the maximum similarity value: \\
    \hspace{0.5cm} $s_{\max}, (i_{\max}, j_{\max}) \leftarrow \max(\mathbf{S})$\;
    
    \If{$s_{\max} < T_h$ \textbf{or} $|\mathbf{C}_{\text{remaining}}| = 0$ \textbf{or} $|\mathbf{R}_{\text{remaining}}| = 0$}{
        \textbf{break}\;
    }
    
    Find all pairs with $s_{\max}$: \\
    \hspace{0.5cm} $\mathbf{P} \leftarrow \{(i, j) \mid S_{ij} = s_{\max}\}$\;
    
    Select one pair randomly: \\
    \hspace{0.5cm} $(i^*, j^*) \leftarrow$ random choice from $\mathbf{P}$\;
    
    Add the matched pair to $\mathbf{M}$: \\
    \hspace{0.5cm} $\mathbf{M} \leftarrow \mathbf{M} \cup \{(\mathbf{R}_{\text{remaining}}[i^*], \mathbf{C}_{\text{remaining}}[j^*])\}$\;
    
    Remove the matched pair from $\mathbf{R}_{\text{remaining}}$ and $\mathbf{C}_{\text{remaining}}$: \\
    \hspace{0.5cm} $\mathbf{R}_{\text{remaining}} \leftarrow \mathbf{R}_{\text{remaining}} \setminus \{\mathbf{R}_{\text{remaining}}[i^*]\}$\;
    \hspace{0.5cm} $\mathbf{C}_{\text{remaining}} \leftarrow \mathbf{C}_{\text{remaining}} \setminus \{\mathbf{C}_{\text{remaining}}[j^*]\}$\;
    
    Remove the corresponding row and column from $\mathbf{S}$: \\
    \hspace{0.5cm} $\mathbf{S} \leftarrow \mathbf{S} \setminus \text{row}(i^*)$\;
    \hspace{0.5cm} $\mathbf{S} \leftarrow \mathbf{S} \setminus \text{column}(j^*)$\;
    
    Iteration $\leftarrow$ Iteration $+ 1$\;
}

Create the result dictionary: \\
\hspace{0.5cm} $\text{result} \leftarrow \{\text{"matched\_features"}: \mathbf{M}, \text{"remaining\_reference\_features"}: \mathbf{R}_{\text{remaining}}, \text{"remaining\_candidate\_features"}: \mathbf{C}_{\text{remaining}}\}$\;
\Return result\;
\end{algorithm}

\section{Experimental Design}
\label{app:hyperparams}

\subsection{Hyperparameters}
We present all our experimental hyperparameters for both generation and evaluation task in Table \ref{tab:hyperparameters} in Appendix. We use a fixed seed and a temperature value of 0.0 across all experiments to ensure the outputs are deterministic and reproducible.

\begin{table}[!htb]
\centering
\caption{Hyperparameters for experiment}
\label{tab:hyperparameters}
\resizebox{\textwidth}{!}{%
\begin{tabular}{@{}lccclc@{}}
\toprule
\multicolumn{1}{c}{\textbf{Models}} &
  \textbf{Seed} &
  \textbf{Temperature} &
  \textbf{Max Token} &
  \multicolumn{1}{c}{\textbf{Message Format}} &
  \textbf{Response Format} \\ \midrule
\textbf{LLaMa-3-70B-Instruct (as generator)} &
  42 &
  0.0 &
  1000 &
  \begin{tabular}[c]{@{}l@{}}\{"role": "system", "content": system\_message\}\\ \{"role": "user", "content": user\_query\}\end{tabular} &
  Default \\ \\ 
\textbf{GPT-4o (as generator)} &
  42 &
  0.0 &
  1000 &
  \begin{tabular}[c]{@{}l@{}}\{"role": "system", "content": system\_message\}\\ \{"role": "user", "content": user\_query\}\end{tabular} &
  Default \\ \\
\textbf{GPT-4o (as evaluator)} &
  42 &
  0.0 &
  1000 &
  \begin{tabular}[c]{@{}l@{}}\{"role": "system", "content": system\_message\}\\ \{"role": "user", "content": user\_query\}\end{tabular} &
  JSON \\ \\ 
\textbf{Trial2Vec / TrialBERT (as evaluator)} &
  Default &
  Default &
  Default &
  Default &
  Default \\ \bottomrule
\end{tabular}%
}
\end{table}
\subsection{Computational Resources used}
We spent around \$120 throughout all of our experiments (both generation and evaluation in zero-shot and three-shot settings) using GPT-4o models. Besides that, we used around 150 compute units from Google Colab for GPU computations. We used  NVIDIA Ampere A100 and NVIDIA T4 GPUs for local inference tasks to calculate BERT scores and other experiments.

\section{Prompts}
\label{app:prompts}

\subsection{Generation Prompt: Zero-shot} 
\begin{figure}[!htb]
    \centering
    \includegraphics[width=\linewidth]{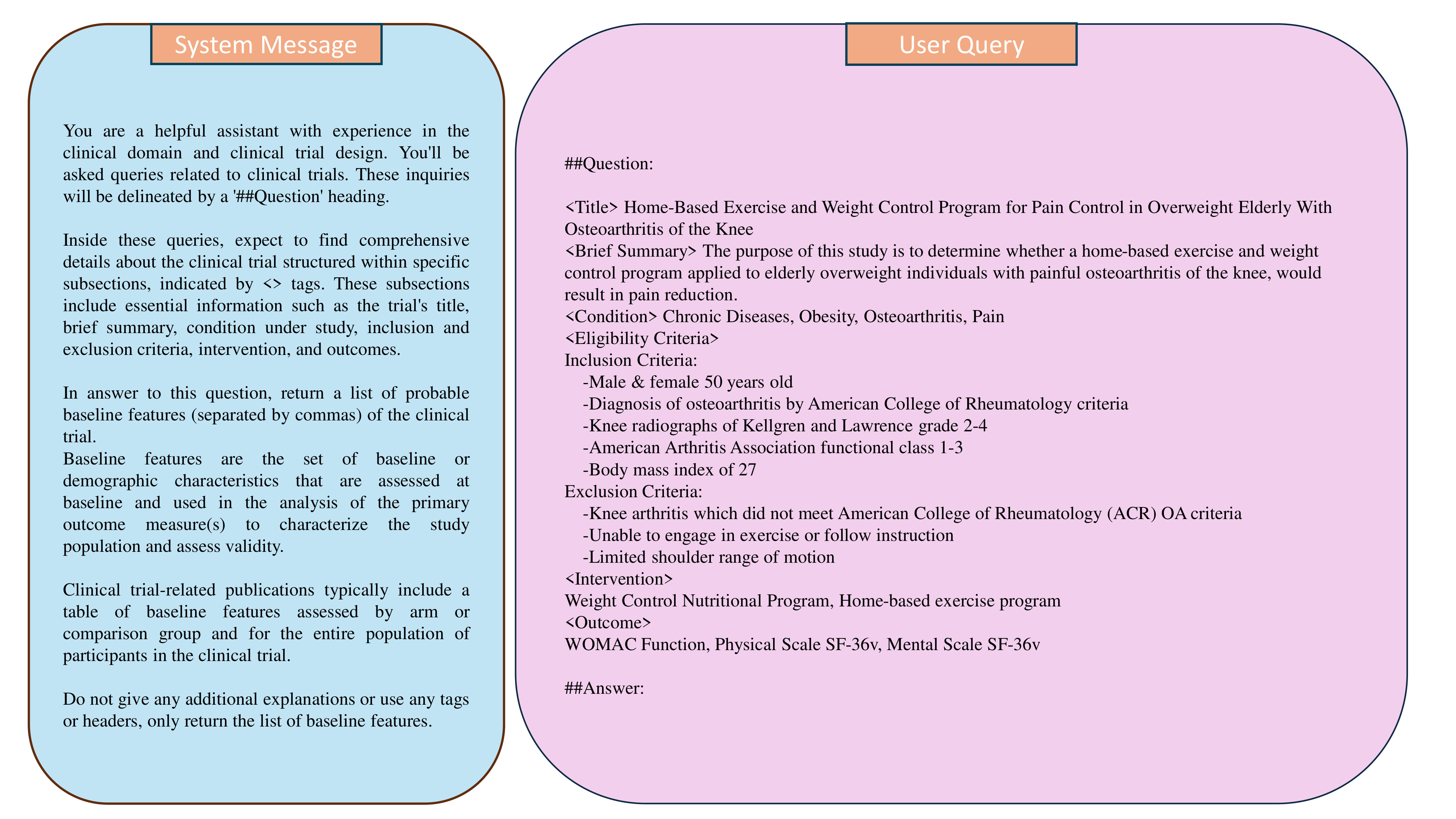}
    \caption{Full Prompt for Generation Task in Zero-Shot setting}
    \label{fig:fullprompt_gen_zs}
\end{figure}

Figure \ref{fig:fullprompt_gen_zs} illustrates the full prompt used to generate LLM responses (i.e., baseline features) in a zero-shot setting. The system message includes detailed instructions for the LLM, specifying the format and structure of the user query. Following this, the user query provides the trial information as context, serving as the question for the LLM.

\subsection{Generation Prompt: Three-shot}

\begin{figure}[!htb]
    \centering
    \includegraphics[width=\linewidth]{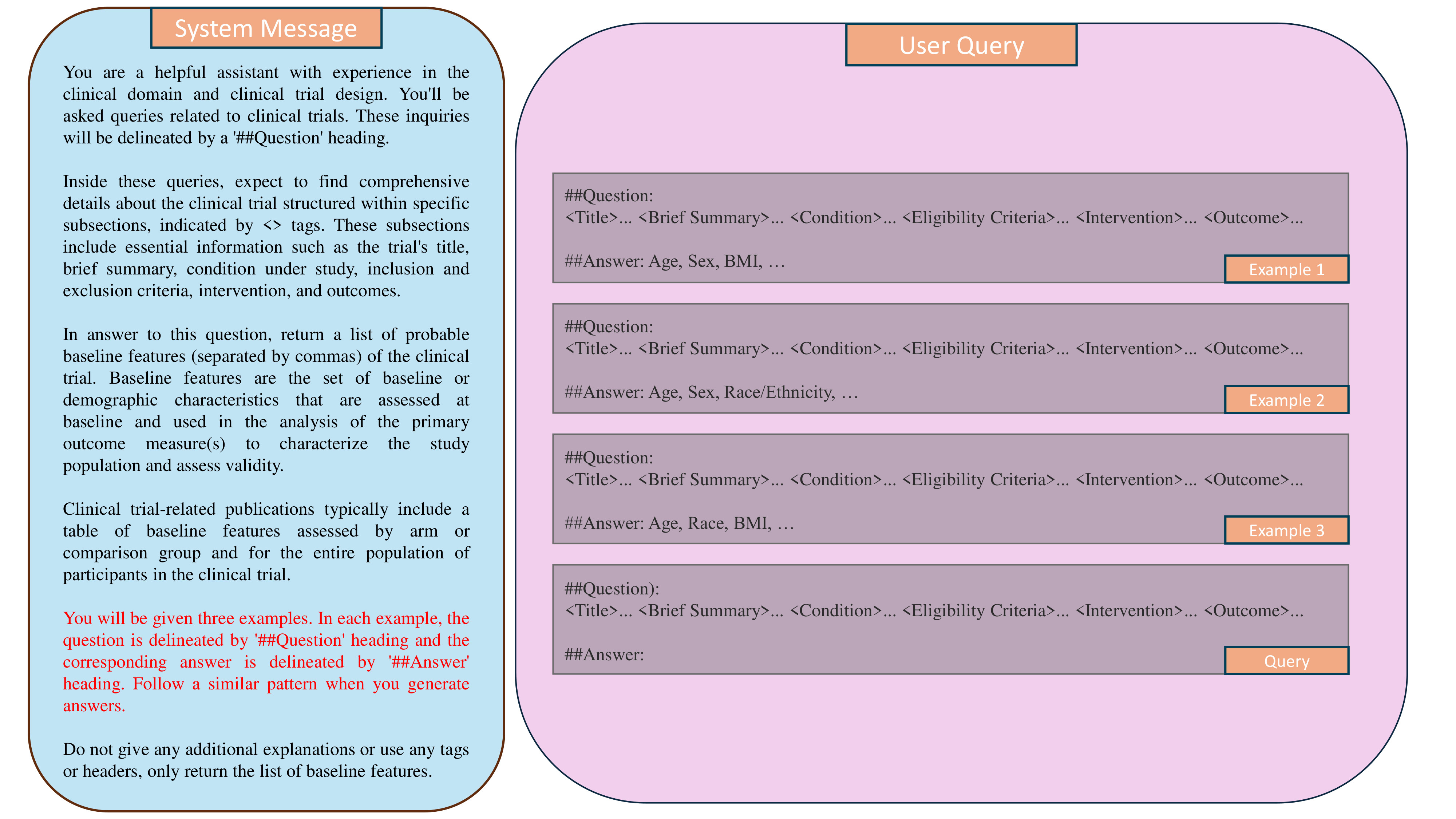}
    \caption{Full Prompt for Generation Task in Three-Shot setting}
    \label{fig:fullprompt_gen_ts}
\end{figure}

Figure \ref{fig:fullprompt_gen_ts} shows the complete prompt used to generate LLM responses (i.e., baseline features) in a three-shot setting. The system message contains detailed instructions for the LLM, including the format and structure of the user query and instructions to expect three examples with their corresponding answers. Next, the user query provides example trial information and their answers as additional context, followed by the actual trial information serving as the question for the LLM.

\subsection{Evaluation Prompt}

\begin{figure}[!htb]
    \centering
    \includegraphics[width=\linewidth]{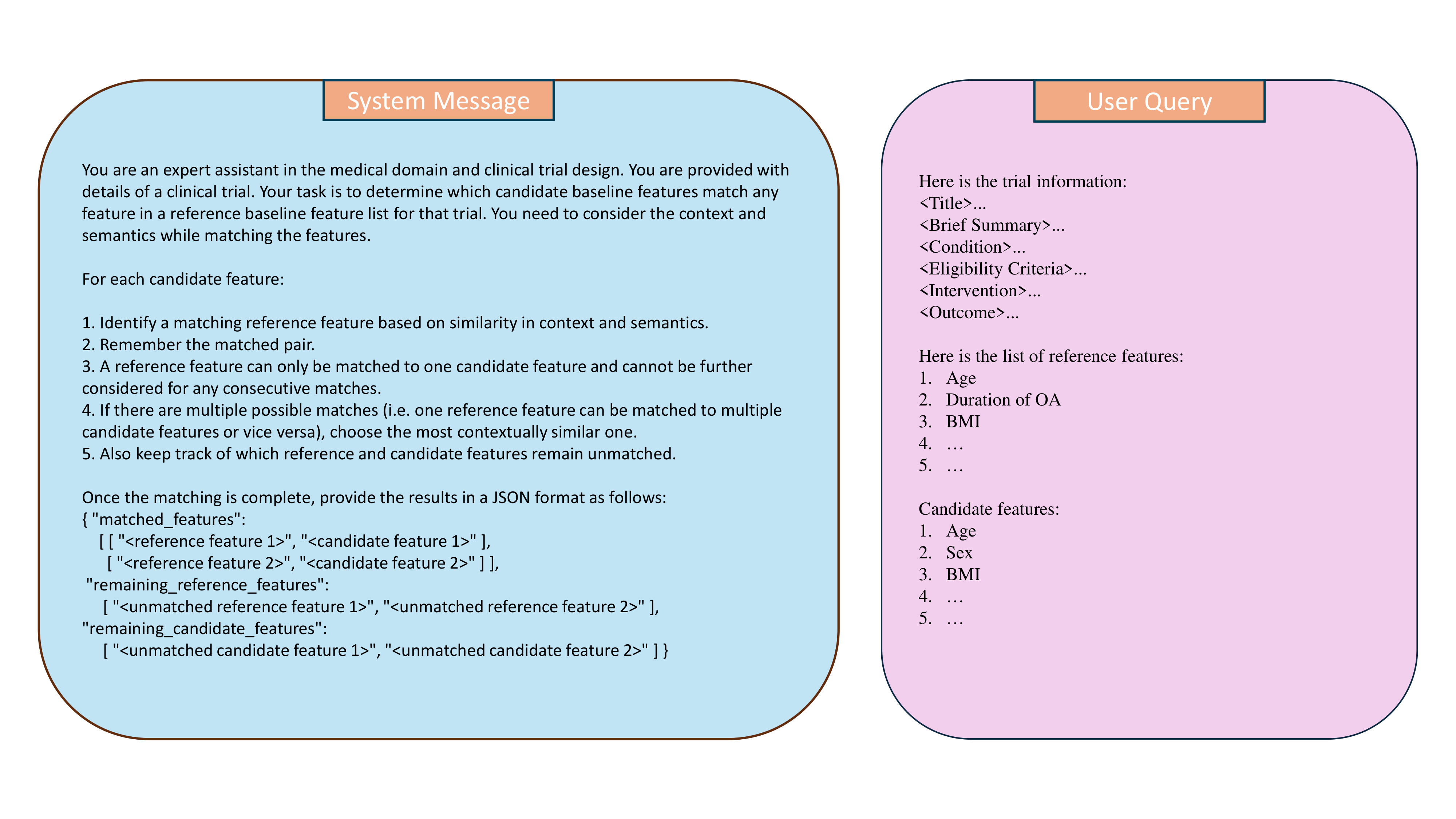}
    \caption{Full Prompt for Evaluation Task}
    \label{fig:fullprompt_eval}
\end{figure}

Figure \ref{fig:fullprompt_eval} displays the complete prompt used to evaluate LLM responses (i.e., candidate features) against a set of reference baseline features. The system message provides detailed instructions for the LLM on how to perform the matching and how to return the response in JSON format. Following this, the user query includes corresponding trial information, along with the list of reference features and candidate features, which serve as the question for the LLM to evaluate.

\section{Additional Experimental Results and Discussion}
\label{app:experiments}

\subsection{BERT Score Comparisons at different thresholds}

For BERT Score, we evaluated four different threshold values $T_h = \{0.6, 0.7, 0.8, 0.9\}$ to be used as similarity thresholds in the similarity matrix (see Appendix A). Figures \ref{fig:ct_pub_plots} and \ref{fig:ct_repo_plots} show the mean precision, recall, and F1 scores for the CT-Pub and CT-Repo datasets, respectively.

\begin{figure}[tb]
    \centering
    \includegraphics[width=\textwidth]{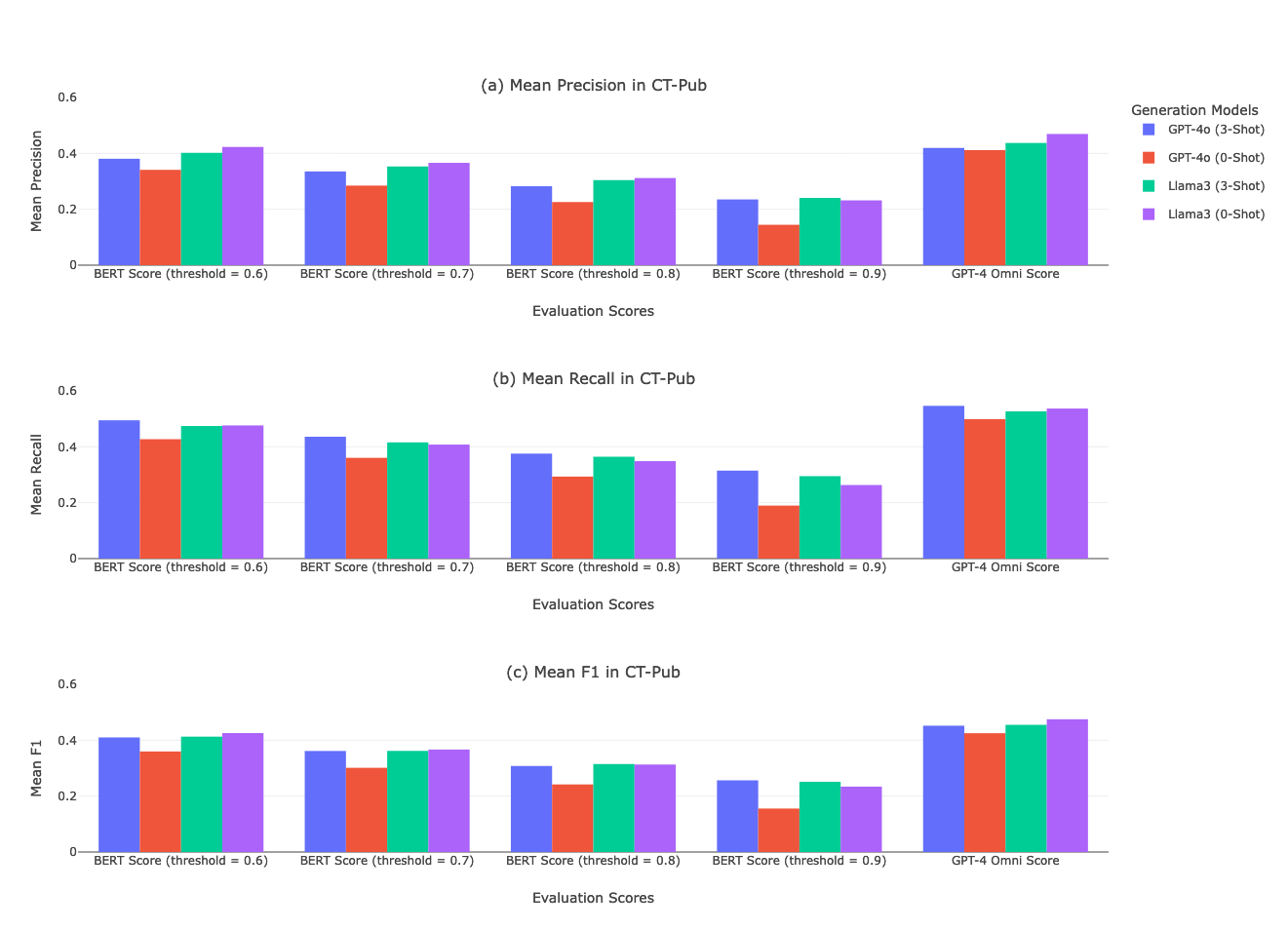}
    \caption{Performance Comparison for CT-Pub dataset}
    \label{fig:ct_pub_plots}
\end{figure}

\begin{figure}[tb]
    \centering
    \includegraphics[width=\textwidth]{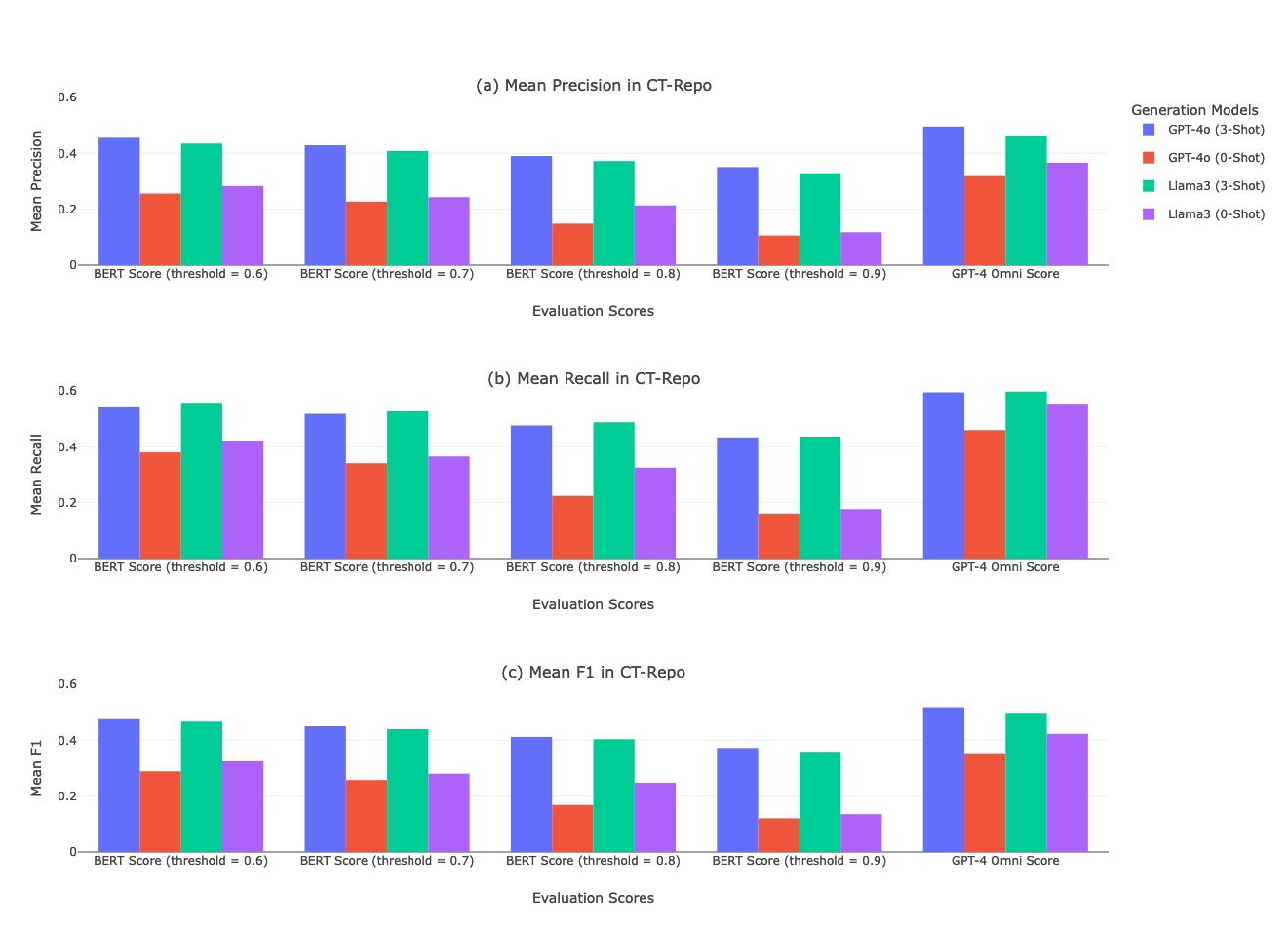}
    \caption{Performance Comparison for CT-Repo dataset}
    \label{fig:ct_repo_plots}
\end{figure}

Increasing the threshold makes the matching more stringent, generally leading to lower precision and recall. Conversely, setting a lower threshold allows more invalid pairs to be matched, indicating a tradeoff. After reviewing several examples and considering domain experts' recommendations, we suggest using a threshold value of 0.7, which balances strictness and accuracy in evaluation.

\subsection{Human Evaluation of GPT-4o's evaluation}
\begin{table}[!htb]
\centering
\caption{Mean of Cohen's Kappa Score for each evaluator pair across all 100 CT-Pub studies}
\label{tab:human-eval}
\resizebox{0.5\textwidth}{!}{%
\begin{tabular}{@{}lc@{}}
\toprule
\textbf{Evaluator Pair} & \textbf{Mean Kappa Score} \\ \midrule
Human 1 and Human 2     & 0.870561                  \\
Human 1 and Human 3     & 0.832767                  \\
Human 2 and Human 3     & 0.831810                  \\
Human 1 and GPT-4o      & 0.847636                  \\
Human 2 and GPT-4o      & 0.816234                  \\
Human 3 and GPT-4o      & 0.783869                  \\ \bottomrule
\end{tabular}%
}
\end{table}
To assess GPT-4o's accuracy as an evaluator, we engaged clinical domain experts to identify matched pairs for 100 CT studies in the CT-Pub dataset. Focusing on GPT-4o's three-shot candidate responses, the experts used the same information and criteria as GPT-4o. We developed a web tool to collect and store their responses. We then compared the responses for the matched pairs from the human evaluators and GPT-4o, creating an inter-rater agreement table and calculating pairwise Cohen's Kappa statistics. Cohen's Kappa measures the agreement level between two raters classifying items into categories. Our findings, presented in Table \ref{tab:human-eval}, show high agreement between the human evaluators and GPT-4o, underscoring GPT-4o's reliability in identifying nuanced feature similarities. The relevant code is available in the GitHub.

\subsection{Sample evaluation responses from GPT-4o, BERT models and human evaluators }
\begin{figure}[tb]
    \centering
    \includegraphics[width=\linewidth]{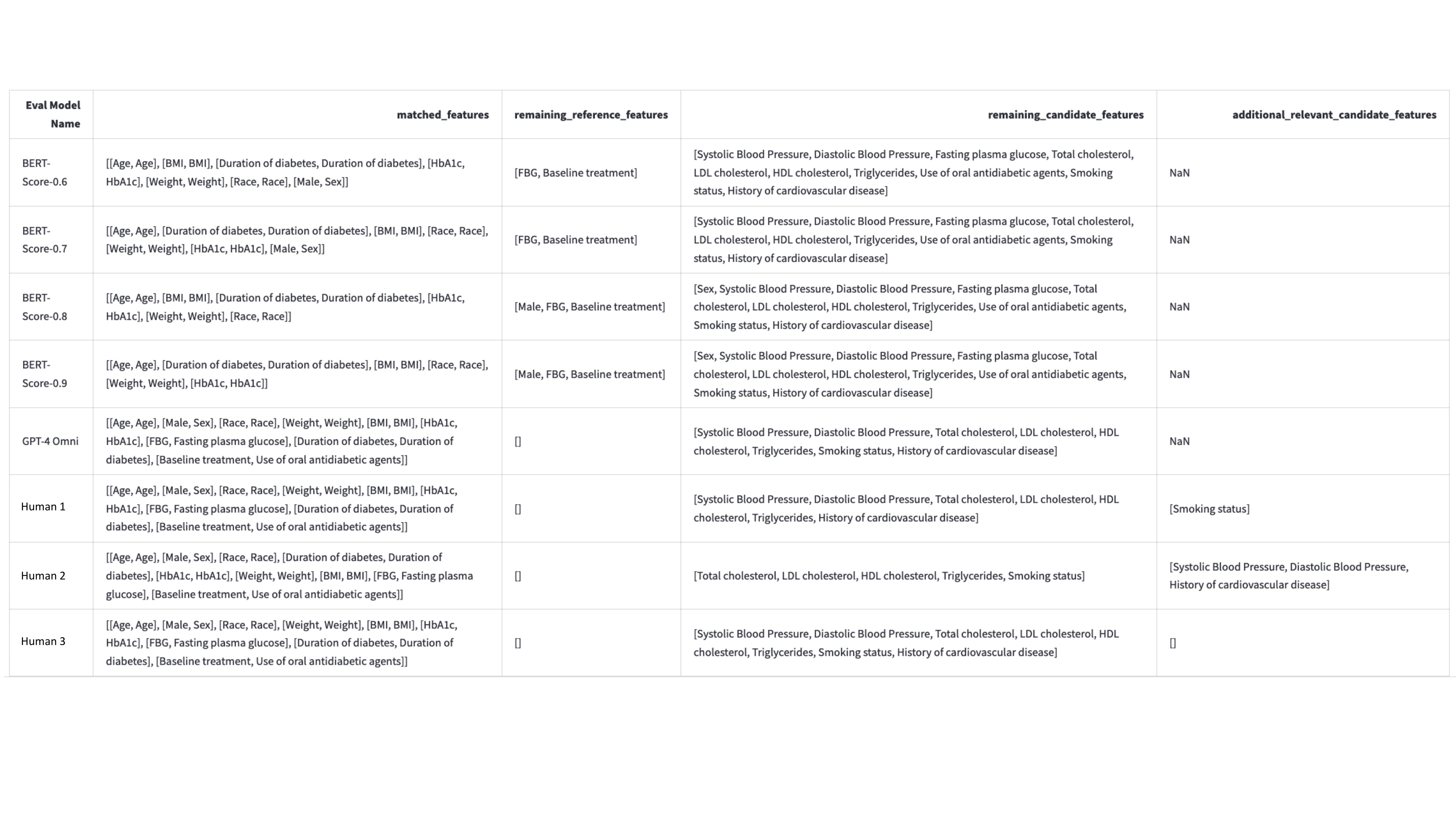}
    \caption{Sample Evaluation Responses}
    \label{fig:eval_responses}
\end{figure}

We present sample evaluation responses, including the list of matched pairs, remaining reference features, remaining candidate features, and additional relevant candidate features in Figure \ref{fig:eval_responses}. The figure presents responses from various evaluation models such as BERT-Score with different threshold values and GPT-4 Omni, as well as three human evaluators. Each row in the figure details the specific features that were matched, those that remained unmatched, and any additional relevant features identified by the corresponding evaluator.

\subsection{Tool Web-UI}

\begin{figure}[!htb]
    \centering
    \includegraphics[width=\linewidth]{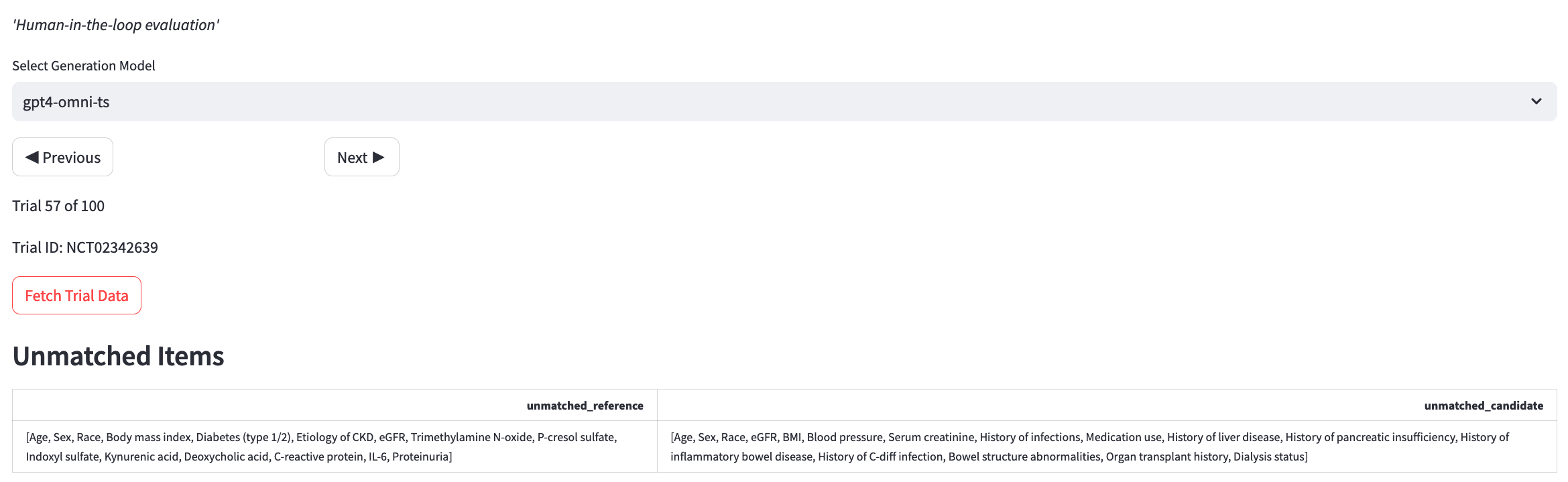}
    \caption{Web Tool UI - Trial Number and Unmatched Items}
    \label{fig:wtu-1}
\end{figure}

\begin{figure}[!htb]
    \centering
    \includegraphics[width=\linewidth]{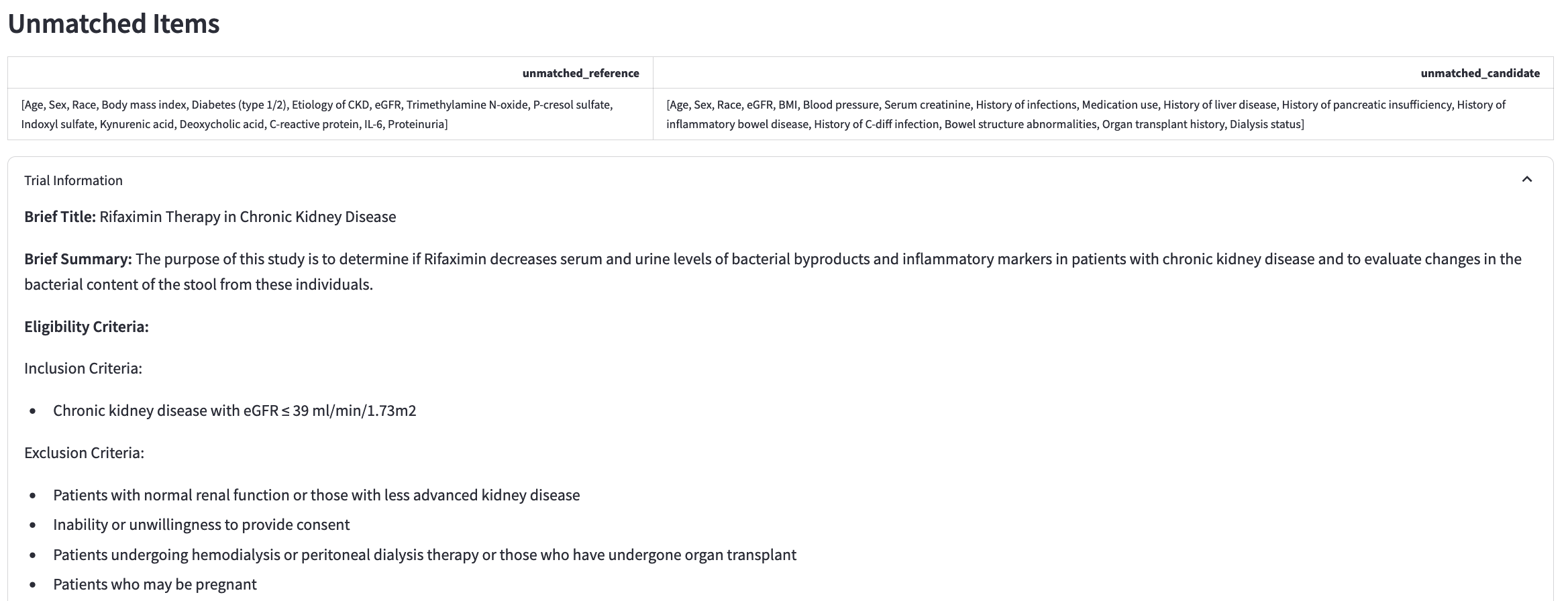}
    \caption{Web Tool UI - Trial Information}
    \label{fig:wtu-2}
\end{figure}

\begin{figure}[!htb]
    \centering
    \includegraphics[width=\linewidth]{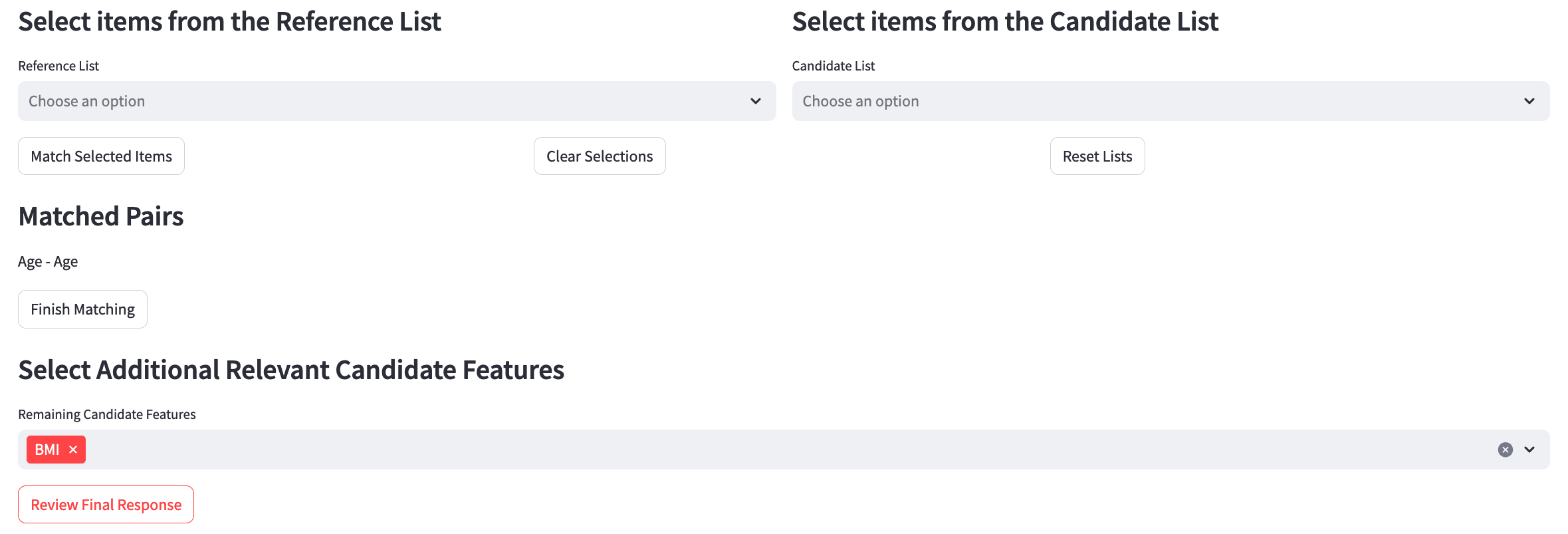}
    \caption{Web Tool UI - Matching}
    \label{fig:wtu-3}
\end{figure}

\begin{figure}[!htb]
    \centering
    \includegraphics[width=\linewidth]{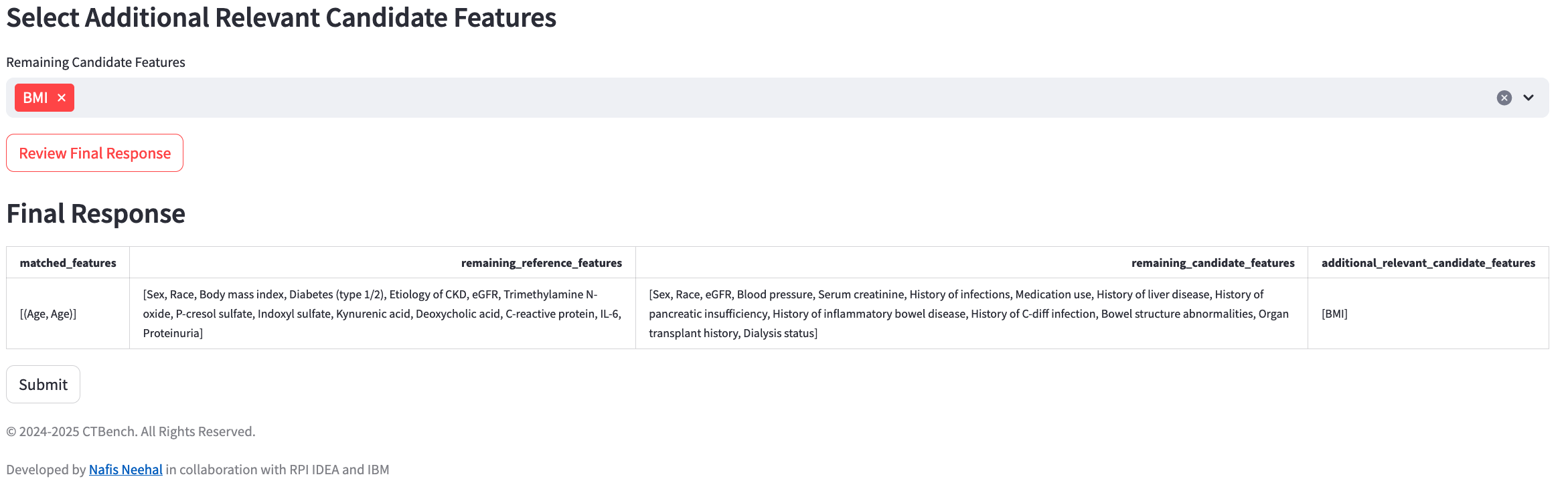}
    \caption{Web Tool UI - Final Response Review}
    \label{fig:wtu-4}
\end{figure}

We provide views of different components of our UI for the web tool to collect responses from each human evaluators. Each of are presented in Figures \ref{fig:wtu-1}, \ref{fig:wtu-2}, \ref{fig:wtu-3} and \ref{fig:wtu-4}.

\end{document}